\pdfoutput=1

\documentclass[11pt]{article}

\usepackage{eacl}

\usepackage{times}
\usepackage{latexsym}

\usepackage[T1]{fontenc}

\usepackage[utf8]{inputenc}

\usepackage{microtype}

\usepackage{inconsolata}

\usepackage{amssymb}
\usepackage{tikz}
\usetikzlibrary{calc, backgrounds, positioning}
\usetikzlibrary{decorations.pathreplacing}
\usepackage{amsmath}
\usepackage{stmaryrd}
\usepackage{url}
\usepackage{booktabs}       
\usepackage{multirow}
\usepackage{paralist}

\usepackage{color}

\title{MAPLE: Micro Analysis of Pairwise Language Evolution for Few-Shot Claim Verification}

\author{Xia Zeng, Arkaitz Zubiaga \\
  Queen Mary University of London \\
  \texttt{\{x.zeng,a.zubiaga\}@qmul.ac.uk} \\
}
\begin{document}
\maketitle

\begin{abstract}
Claim verification is an essential step in the automated fact-checking pipeline which assesses the veracity of a claim against a piece of evidence. In this work, we explore the potential of few-shot claim verification, where only very limited data is available for supervision. We propose MAPLE (Micro Analysis of Pairwise Language Evolution), a pioneering approach that explores the alignment between a claim and its evidence with a small seq2seq model and a novel semantic measure. Its innovative utilization of micro language evolution path leverages unlabelled pairwise data to facilitate claim verification while imposing low demand on data annotations and computing resources. MAPLE demonstrates significant performance improvements over SOTA baselines SEED, PET and LLaMA 2 across three fact-checking datasets: FEVER, Climate FEVER, and SciFact. Data and code are available \href{https://github.com/XiaZeng0223/MAPLE}{here}.

\end{abstract}

\begin{figure*}[t]
    \centering
    \includegraphics[width=\textwidth]{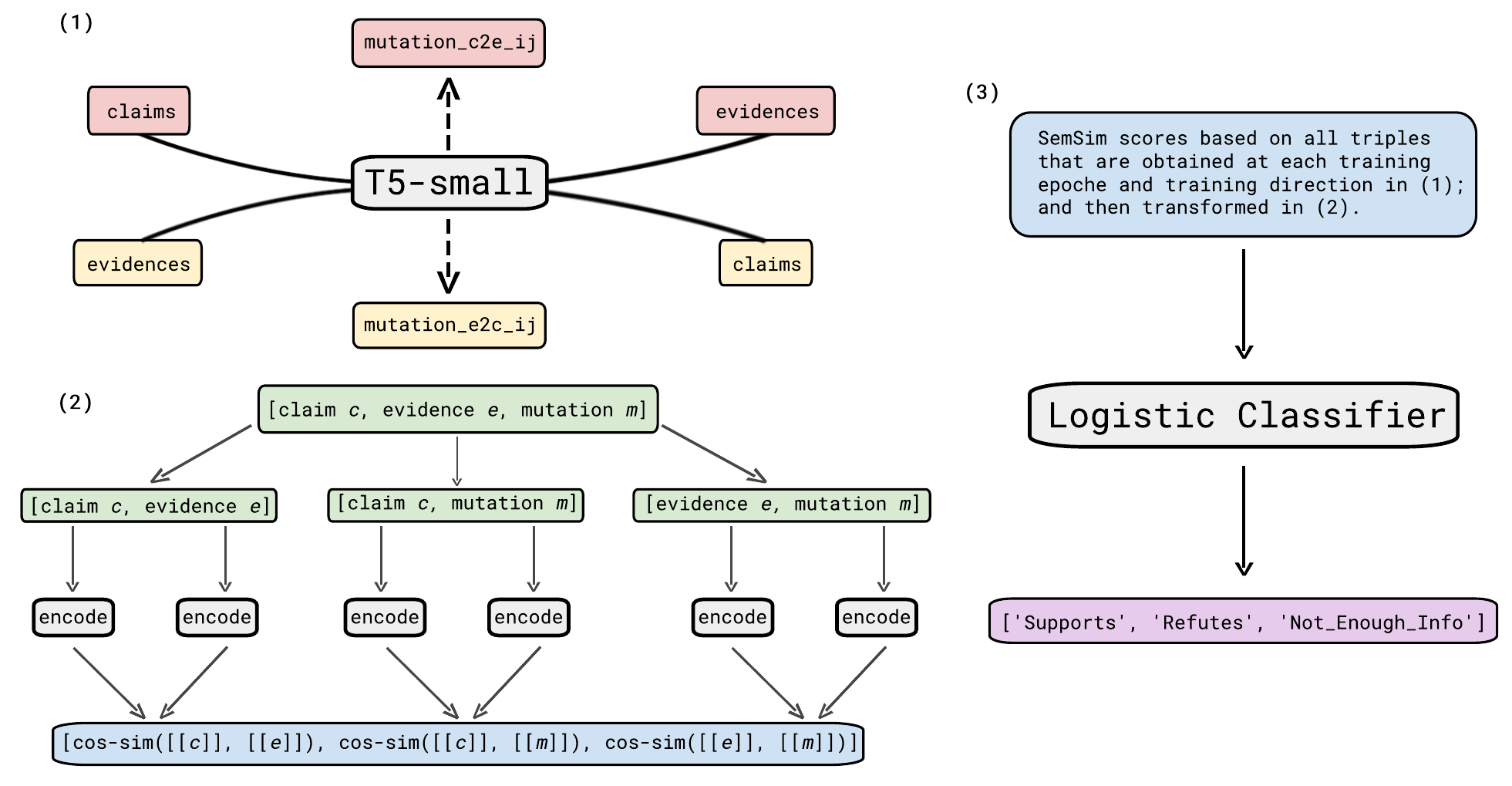}
    \caption{MAPLE for claim verification. \textbf{(1) In-domain seq2seq training.} With LoRA, a T5-small model is trained on claim-to-evidence task for $e$ epochs using the $d$ unlabelled claim-evidence pairs from the data pool. At the end of each training epoch $j$, model inference is performed on each instance $i$ to generate a mutation $mutation\_c2e\_i$. This process is repeated on evidence-to-claim setting. In total this step produces $2*d*e$ triples that consist of a claim $c$, an associated piece of evidence $e$ and a generated mutation $m$.
    \textbf{(2) SemSim transformation.} Each triple is grouped into three pairs including claim-evidence pair $c-e$, claim-mutation pair $c-m$ and evidence-mutation pair $e-m$. `Semsim' scores are obtained for each pair by calculating the cosine similarity score based on corresponding sentence embeddings.
    \textbf{(3) Logistic classifier training with few-shot labelled data.} A logistic classifier is trained on labelled data where the transformed `SemSim' scores are used input features to predict veracity labels.}
    \label{figure:ill}
\end{figure*}

\section{Introduction}
The proliferation of misinformation and fake news has become a significant concern in today's information landscape. Fact-checking has emerged as a crucial task to combat the spread of false information \cite{thorne_automated_2018, kotonya_explainable_2020, nakov_automated_2021, zeng_automated_2021, guo_survey_2022}. A body of natural language processing (NLP) research has investigated the task of claim verification: determining the veracity of a claim based on retrieved evidence. It is often addressed in a Natural Language Inference (NLI) fashion, namely making predictions on the claim with reference to evidence out of three candidate labels: `SUPPORTS', `REFUTES', and `NOT\_ENOUGH\_INFO'. While the majority of previous work tackles the problem with fully supervised methods \cite{li_paragraph-level_2021, zeng_qmul-sds_2021, zhang_abstract_2021, wadden_multivers_2022, rana_rerrfact_2022, rana_levirank_2022}, deploying these methods face practicality issues. Emerging domains of misinformation often involve novel claims, limiting the availability of relevant labeled data. Fact-checkers often need to evaluate claims with time constraints, limiting the time allowed for conducting extensive fine-tuning of pretrained language models (PLMs). Hence, performing claim verification in few-shot scenarios is of particular importance in the real-world combat of misinformation. 

The current state-of-the-art (SOTA) methods for few-shot claim verification are Semantic Embedding Element-wise Difference (SEED) \citep{zeng_aggregating_2022} and Pattern Exploiting Training (PET) \citep{schick_exploiting_2021, schick_its_2021}. However, their few-shot performance relies on the use of NLI-trained PLMs, limiting their applicability to only cases where NLI data and NLI-trained PLMs are available, excluding scenarios such as low-resource languages. Moreover, these methods excel when the data is similar to NLI data but struggle when dealing with dissimilar data. In contrast, we propose to embrace the potential of leveraging unlabeled data, which is more readily available in a fact-checking pipeline, to enhance few-shot claim verification.

An alternative strand of research in the realm of general few-shot classification advocates for generative Large Language Models (LLMs) endowed with billions of parameters, exemplified by models like GPT-4 \cite{openai2023gpt4} and LLaMA 2 \cite{touvron2023llama}. These models demonstrate impressive few-shot performance, though introducing a reliance on advanced computational resources and prolonged inference times. In contrast, our work challenges this paradigm by demonstrating that smaller models, such as T5-small \cite{raffel_exploring_2020}, possess the inherent capability to excel in few-shot learning scenarios. Leveraging unlabeled data and advanced semantic measures, our approach underscores the efficacy of compact models in achieving effective and robust few-shot performance without the need for extensive computational resources.

We present MAPLE (Micro Analysis of Pairwise Language Evolution), a novel approach designed for few-shot claim verification. MAPLE innovatively builds upon the concept of language transition\footnote{In this paper, we distinguish between claim language and evidence language, treating them as distinct languages as they may differ in formality, length, or even depth. In real-world scenarios, checkworthy claims often emanate from more informal settings, such as social media platforms. On the other hand, evidences typically come from formal and reputable sources such as research papers and Wikipedia, marked by a concise, informative, and professional style. For concrete examples, please see the data samples in Appendix \ref{appendix: datasets}.}, scrutinizing the semantic shift that occurs as a sequence-to-sequence model learns to generate a target sequence from a given input sequence. In this paper, such language transition from the input sequence to the output sequence over the training epochs is referred to as pairwise language evolution. By intricately capturing and harnessing this pairwise language evolution, MAPLE aims to facilitate accurate predictions even in scenarios with minimal labeled data. Our key novel contributions include:

\begin{compactitem}
\item We introduce MAPLE, an innovative approach that leverages unlabeled data for enhancing few-shot claim verification. While building MAPLE, we also propose `SemSim' as an NLG evaluation metric that focuses on semantic similarity.
\item We perform a pioneering exploration of the language transition convergence process during seq2seq model training.
\item We conduct comprehensive experiments on four dataset configurations, facilitating a direct comparison with established SOTA methods, namely SEED, PET, and LLaMA 2.
\end{compactitem}

\section{Related Work}

\subsection{Few-Shot Learning for Claim Verification}

One initial attempt in this direction was made by \citet{lee_towards_2021}, who proposed a perplexity-based approach using language models. However, this approach is restricted to binary classification and underperforms recent advancements. In contrast, \citet{zeng_aggregating_2022} introduced SEED, a method that calculates PLM-based pairwise semantic differences between claims and associated evidence. By deriving representative class vectors from these differences, SEED offers an efficient solution for few-shot claim verification and serves as one of our baseline models.

Another competitive training procedure for few-shot learning is PET \citep{schick_exploiting_2021, schick_its_2021}. PET reformulates classification tasks into cloze tasks using templates. By calculating the probability of candidate tokens filling the placeholder [mask] position with an PLM, PET maps it to a preconfigured label. PET has demonstrated its few-shot capabilities in various NLP benchmarks, including claim verification \citep{zeng_active_2023}.\footnote{In \citet{zeng_active_2023}, we proposed ActivePETs as an active learning method, which focuses on data annotation prioritisation. Despite both tackling claim verification, ActivePETs is not a fair comparison with MAPLE, which is a few-shot classification method focused on achieving better performance with robustness to random sampling.} Though SEED and PET have been proposed as methods for few-shot claim verification, the evaluation datasets they used differ from each other. To address this gap and broaden the evaluation, we conduct experiments on four dataset configurations, allowing for a direct comparison.

When addressing claim verification, both SEED and PET heavily rely on PLMs trained on NLI, which brings several limitations. Firstly, they face challenges when dealing with data that significantly differs from general NLI datasets, such as cases where the domain is highly technical and different from general NLI data pairs and/or the evidence consists of large paragraphs rather than single sentences. Additionally, their reliance on NLI-trained models restricts their applicability to languages for which NLI datasets and corresponding PLMs are available, excluding their use in low-resource languages. Moreover, Our proposed model MAPLE does not rely on NLI-trained models but instead utilizes unlabelled claim-evidence pairs which could be abundant and useful for domain adaptation.

In addition, recent advancements in generative LLMs with multi-billion parameters have showcased impressive few-shot capabilities. However, closed-source pioneering models, including GPT-3.5 and GPT-4, present reproducibility challenges with their behavior changing over time \cite{chen2023chatgpts}. In this study, we prioritize open-source solutions, with a particular focus on LLaMA 2, a recent model that surpasses existing open-source alternatives across various benchmarks \cite{touvron2023llama}. The primary drawback of these approaches lies in their requirement for advanced computational infrastructure, a substantial computational budget, and extended inference times. MAPLE tackles these constraints by utilizing parameter-efficient models, aiming to improve both resource and runtime efficiency.

\subsection{Natural Language Generation (NLG) Metrics}
NLG evaluation metrics play a crucial role in evaluating the quality of generated texts. Classic metrics such as BLEU (Bilingual Evaluation Understudy) \citep{papineni_bleu_2002}, ROUGE (Recall-Oriented Understudy for Gisting Evaluation) \citep{lin_rouge_2004}, and METEOR (Metric for Evaluation of Translation with Explicit ORdering) \citep{banerjee_meteor_2005} remain as the most widely used metrics. They address the evaluation as a matching task, quantifying n-gram overlap with recall, precision and F-score and providing lexical-level evaluations. 
Recent advancements include SacreBLEU \citep{post_call_2018}, which enhances reproducibility, tokenization support, and ease of statistical significance reporting. In contrast, BLEURT (Bilingual Evaluation Understudy with Representations from Transformers) \citep{sellam_bleurt_2020} advances semantic-level evaluations and treats evaluation as a regression task using PLMs. Another metric, BARTScore \citep{yuan_bartscore_2021}, approaches evaluation as a text generation task for LLMs, calculating the BARTScore as the weighted log probability of one text given another text.

Given our primary interest in the semantic shift during pairwise language evolution, we propose `SemSim' as an alternative metric to evaluate NLG performance.

\subsection{Understanding Language Evolution}

Language evolution has been the subject of several theories, including biological evolution, learning, and cultural evolution \citep{lekvam_agent-based_2014}. Studies conducted in laboratory settings have explored the intricate nature of various phenomena, offering valuable insights into the emergence of language \citep{scott-phillips_language_2010}.

Researchers have focused on modeling evolution within language families to identify patterns in phonetic features across observed languages \cite{nouri_modeling_2016}. Computational research has also introduced tools such as language evolution simulators, examining word-level evolution within language families \citep{ciobanu_simulating_2018}, and realistic geographic environments to simulate language and linguistic feature development over time \citep{kapur_modeling_2020}. These studies tackle various related issues for historical linguistics, areal linguistics, and linguistic typology.

While language evolution research often adopts a macro and historical perspective, this paper engages in micro-level analysis, i.e. asking ``what path does it take for a piece of text to migrate into another piece”. Interestingly, the convergence process during seq2seq training simulates such a path of evolving or transitioning language. In our work, we investigate language transition across seq2seq training epochs and further utilize it to conduct pairwise classification.

\section{Methodology}

Traditionally, generative models are often used in classification tasks by generating corresponding labels given input sentences \citep{pradeep_scientific_2021}. However, such an approach does not fully exploit the potential of generative models on tasks such as claim verification. In this section, we present the MAPLE method and its application to few-shot claim verification. 

The intuition of MAPLE is that sentence pairs of various relationships bring diverse learning challenges to the seq2seq generation task. As the data difficulty is reflected in the seq2seq training process, such learning difficulty associated with each sample could be further transformed into various signs to indicate the relationship within a sentence pair. We explore such potential to be leveraged for effective claim verification, where the goal is to determine the veracity of a claim based on its relationship with the provided evidence. MAPLE consists of three steps, as illustrated in Figure \ref{figure:ill}.

\paragraph{\textbf{(1) In-domain seq2seq training.}} 
In order to leverage in-domain unlabeled data, i.e. claim-evidence pairs without veracity labels, we perform seq2seq training in two directions: claim-to-evidence and evidence-to-claim. For claim-to-evidence task, a T5-small \cite{raffel_exploring_2020} model is fine-tuned for $e$ epochs using all of the unlabeled claim-evidence pairs from the data pool with a size of $d$. At the end of each training epoch $j$, model inference is performed on each instance $i$ to generate a mutation $mutation\_c2e\_i$. Similarly, another T5-small model is fine-tuned on evidence-to-claim task to generate mutations $mutation\_e2c\_i$ for each training epoch $j$. For computational efficiency, the training is conducted with Low-Rank Adaptation (LoRA) \cite{hu2021lora}, a parameter-efficient training method. In total, this step produces $2*d*e$ triples that consist of a claim $c$, an associated piece of evidence $e$ and a generated mutation $m$.

\paragraph{\textbf{(2) SemSim transformation.}} 
The SemSim transformation aims to transform the generated triples into numeric scores while recording the transition of mutation $m$ during the training process in both claim-to-evidence task and evidence-to-claim task.
Each triple is grouped into three pairs including claim-evidence pair $c-e$, claim-mutation pair $c-m$ and evidence-mutation pair $e-m$. We measure the pairwise similarity with `SemSim' score: first obtains sentence embeddings with model `sentence-transformers/all-mpnet-base-v2' \cite{reimers_sentence-bert_2019}, a sentence transformer model that is trained on over one billion sentences with contrastive training objective; then calculates cosine similarity scores on sentence embeddings for each pair. Each triple is transformed into an array of 3 `SemSim' scores. All triples of a claim-evidence instance are concatenated as features of the instance.

\paragraph{\textbf{(3) Logistic classifier training with few-shot labeled data.}} 
Using $n$-shot labeled data from the labeled data pool of size $3n$,\footnote{For example, 1-shot experiments are conducted on a data pool that includes 3 labeled samples in total, i.e., one instance per class per claim verification task.} i.e. claim-evidence pairs with veracity labels, a logistic classifier is trained. The transformed SemSim scores are used as input features to make predictions on veracity labels.\footnote{Please note that MAPLE differs from data augmentation methods. Data argumentation generates pseudo-data and uses them as additional samples for model training; MAPLE does not treat mutations as additional training samples, but relies on them to obtain input features for logistic classifier training. From a tabular view, typical data augmentation methods generate additional rows but MAPLE operates on columns.}

\section{Experiments}
\label{experiments}
In this section, experiments comparing MAPLE with previous SOTA methods are presented. 

\subsection{Datasets}
We carry out experiments on four dataset configurations using three datasets: FEVER, climate FEVER, and SciFact. The FEVER dataset is the first large-scale fact-checking dataset and has had a significant impact in the field. SciFact and climate FEVER datasets are known to be challenging, technical, and free of synthetic data. Corresponding data samples and label distributions can be found in Appendix \ref{appendix: datasets}.

\paragraph{FEVER}
FEVER \citep{thorne_fever_2018} is a large-scale dataset for automated fact-checking. It contains claims that are manually modified from Wikipedia sentences along with their corresponding Wikipedia evidences. Despite criticisms of its synthetic nature by researchers in the fact-checking domain, it has been widely used also outside of fact-checking. Various NLP benchmarks, such as KILT \cite{petroni_kilt_2021}, include the claim verification task of FEVER to test models' reasoning capabilities. As is common in the general NLP community, we follow the practice of using oracle evidence, skipping the evidence retrieval step. We only use the test set of the original FEVER dataset, as it contains higher-quality data and the quantity is sufficient for few-shot experiments. We reserve 150 instances for each class to form a test set and leave the rest in the train set.

\paragraph{cFEVER}
Climate FEVER \citep{diggelmann_climate-fever_2021} is a challenging, large-scale dataset that consists of claim and evidence pairs related to climate change, along with their veracity labels. Since the dataset does not naturally provide options for setting up retrieval modules, we directly use it for the claim verification task. Similarly, we reserve 150 instances for each class to form a test set and leave the rest in the train set.

\paragraph{SciFact}
SciFact \citep{wadden_fact_2020} provides scientific claims with their veracity labels, along with a collection of scientific paper abstracts, some of which contain rationales to resolve the claims. Additionally, it provides oracle rationales that can be linked to each claim. Unlike FEVER, research on SciFact places strong emphasis on the evidence retrieval module. Hence, we conduct experiments on SciFact with two configurations: SciFact\_oracle and SciFact\_retrieved. The former utilizes oracle evidence provided by the annotations, while the latter uses evidence retrieved by a retrieval model, namely BM25, to retrieve the top 3 abstracts as evidences \citep{wadden_multivers_2022, zeng_active_2023}. We merge the original SciFact train set and dev set and redistribute the data to form a test set that contains 150 instances for each class, using the rest as the train set.

\subsection{Baselines}

\paragraph{SEED}
SEED uses a sentence-transformer model that is trained on NLI tasks.\footnote{Huggingface hub model id `bert-base-nli-mean-tokens' \citep{zeng_aggregating_2022}.}

\paragraph{PET}
PET uses BERT-base fine-tuned on the MNLI dataset.\footnote{Huggingface hub model id `textattack/bert-base-uncased-MNLI'. See performance using alternative model checkpoint in Appendix \ref{appendix: detailed performance comparison}.} It is trained with a batch size of 16, a learning rate of $1e^{-5}$, and training epochs of 3, following previous practice \citep{schick_exploiting_2021, schick_its_2021, zeng_active_2023}.

\paragraph{ LLaMA 2}
LLaMA 2 experiments are conducted on the LLaMA 2 7b chat model.\footnote{Huggingface hub model id `Llama-2-7b-chat-hf'. See performance using alternative model checkpoint in Appendix \ref{appendix: detailed performance comparison}.} Answers are generated by prompting with detailed instructions\footnote{After evaluating several prompts, the subsequent one is employed due to its superior performance.: ``Please perform the task of claim verification: you are given a claim and a piece of evidence, your goal is to classify the pair out of `SUPPORTS', `REFUTES' and `NOT\_ENOUGH\_INFO'. Here are a few examples: claim: {train\_claim\_i} evidence: {train\_evidences\_i} label: {train\_label\_i} What is the label for the following pair out of `SUPPORTS', `REFUTES' and `NOT\_ENOUGH\_INFO'? Answer with the label only. ''} and post-processed to match class labels \footnote{Post-processing primarily includes stripping formatting strings and removing ``label: ''. The remaining responses that do not belong to any of the labels are mapped into the ``NOT\_ENOUGH\_INFO'' class, e.g. responses such as ``?'' and ``Please give me the answer''.}.

\subsection{MAPLE}
In our experiments, MAPLE uses the T5-small model for efficient training.\footnote{Huggingface hub model id `t5-small' \cite{raffel_exploring_2020}.} Training is conducted with LoRA from epoch 0 to epoch 20, using 0.0001 as learning rate, 16 as batch size, 512 as max length, 0.1 as LoRA dropout, 32 as LoRA alpha \cite{hu2021lora} and ``Summarize:'' as the prompt \citep{ramamurthy_is_2023}.

\subsection{Experimental Setup}
Our experimental setup is designed to conduct comprehensive few-shot experiments, where the term `n-shot' refers to the number of samples available per class. As we focus on few-shot performance, our main experiments are conducted on 1-shot, 2-shot, 3-shot, 4-shot and 5-shot settings. To ensure the reliability and generalizability of our findings, each n-shot experiment has been repeated 100 times with sampling seeds ranging from 123 to 223. We present the main results in Section \ref{section:results}. We also present further experiments showing the trend going up to 50 shots in Appendix \ref{appendix: performance within 50 shots}.

\begin{figure*}[htb]
    \centering
    \includegraphics[scale=0.28]{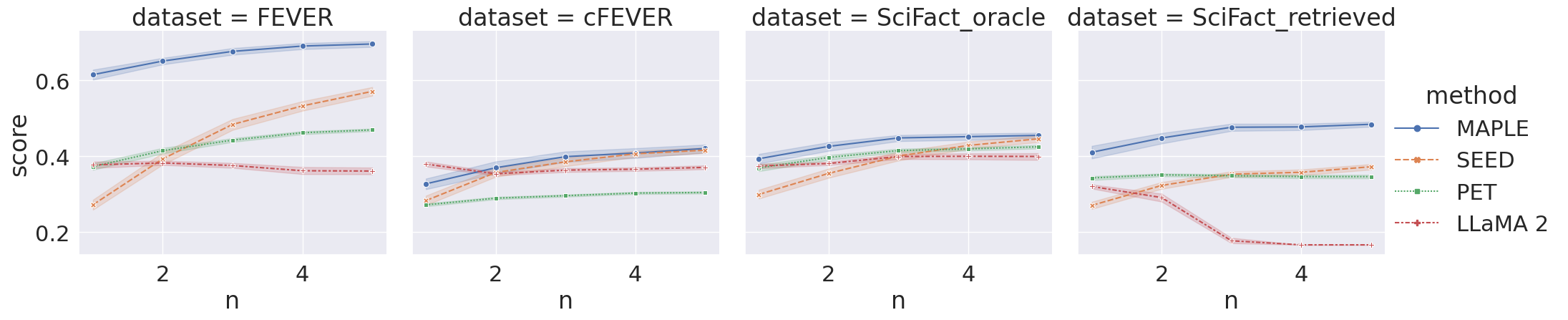}
    \caption{F1 performance within 5 shots.}
    \label{fig:max5}
\end{figure*}

\section{Results}
\label{section:results}
In this section, we present the results of our experiments with a focus on few-shot settings. 

Figure \ref{fig:max5} illustrates the F1 performance within the 5-shot setting.\footnote{Please see detailed classwise performance in Appendix \ref{appendix: MAPLE Classwise Performance within 5 shots}} Across the four dataset configurations, MAPLE shows noticeable performance advantages within the 5-shot setting, validating its effectiveness in few-shot scenarios and robustness across datasets. It achieves this primarily by starting from a high performance point and steadily improving within 5 shots. Although SEED underperforms MAPLE, it showcases strong learning capabilities, and its relatively lower performance is primarily due to a low starting point. Surprisingly, PET and LLaMA 2 perform poorly within the 5-shot range, generally starting low and exhibiting limited learning capabilities.  

On the FEVER dataset, MAPLE demonstrates significant improvements over the baselines. Specifically, MAPLE achieves a very high F1 score over 0.6 at 1 shot, outperforming SEED, PET, and LLaMA 2, which commence at approximately 0.25, 0.37, and 0.38, respectively. Within 5 shots, MAPLE exhibits a steady performance improvement, surpassing an F1 score of 0.7. While SEED and PET also experience notable performance boosts, with SEED approaching just below 0.6 and PET reaching below 0.5, LLaMA 2 encounters a slight performance drop, settling around 0.36. 

On the cFEVER dataset, the performance of all methods exhibits a considerable decrease compared to FEVER, highlighting the challenging nature of the dataset. While MAPLE maintains its leading position overall, the performance margin is narrower. It initiates above 0.3 and achieves scores surpassing 0.4. SEED begins even lower, below 0.3, but manages to surpass 0.4, albeit slightly trailing behind MAPLE. PET encounters greater challenges overall, commencing below SEED and only slightly exceeding 0.3. LLaMA 2 excels initially with a score of 0.38 but experiences a drop to 0.37. 

On the SciFact\_oracle dataset configuration, despite the overall performance being better than cFEVER but worse than FEVER across all methods, MAPLE maintains superiority within 5 shots. It initiates around 0.4 and concludes around 0.45. SEED begins around 0.3 and lags behind MAPLE, while PET starts higher than SEED but lower than MAPLE, failing to surpass them within 5 shots. LLaMA 2 performs comparably to PET, starting at 0.37 and finishing at 0.40. 

On the SciFact\_retrieved dataset configuration, MAPLE demonstrates a slightly better performance compared to SciFact\_oracle, while all baseline methods exhibit a substantial decline in performance compared to SciFact\_oracle. Consequently, MAPLE achieves a larger performance margin. It commences above 0.4 and concludes around 0.5. SEED starts at a very low point, below 0.3, and approaches 0.4 at 5 shots. PET initiates around 0.35 but struggles to learn effectively within 5 shots, resulting in an even lower score. LLaMA 2 starts at 0.32 and 0.29 and experiences a notable drop to 0.18 and 0.17 immediately afterwards.\footnote{Note that the SciFact\_retrieved dataset configuration comprises lengthy instances that may exceed the maximum context length for LLaMA 2. Addressing this issue would necessitate additional techniques.}

In general, LLaMA 2 displays reasonable one-shot performance but shows limited learning capabilities within 5 shots. Despite PET's use of gradient descent to update the parameters of a large language model, this strategy does not yield satisfactory results within the 5-shot range. On the other hand, MAPLE and SEED showcase relatively rapid convergence due to their limited number of trainable parameters. MAPLE stands out with a significantly higher level of performance compared to all baselines overall, demonstrating its capacity to leverage limited data for notable results and effectiveness as a few-shot claim verification model. 

It's crucial to highlight that while most experiments are conducted in oracle settings, real-world claim verification often introduces the challenge of imperfect evidences. Therefore, achieving optimal performance in the SciFact\_retrieved dataset, where evidence is noisy and lengthy, is particularly significant. This accomplishment highlights MAPLE's robustness to noisy and challenging data in realistic fact-checking scenarios.

\section{Ablation Studies}

\begin{figure*}[htb]
    \centering
    \includegraphics[scale=0.27]{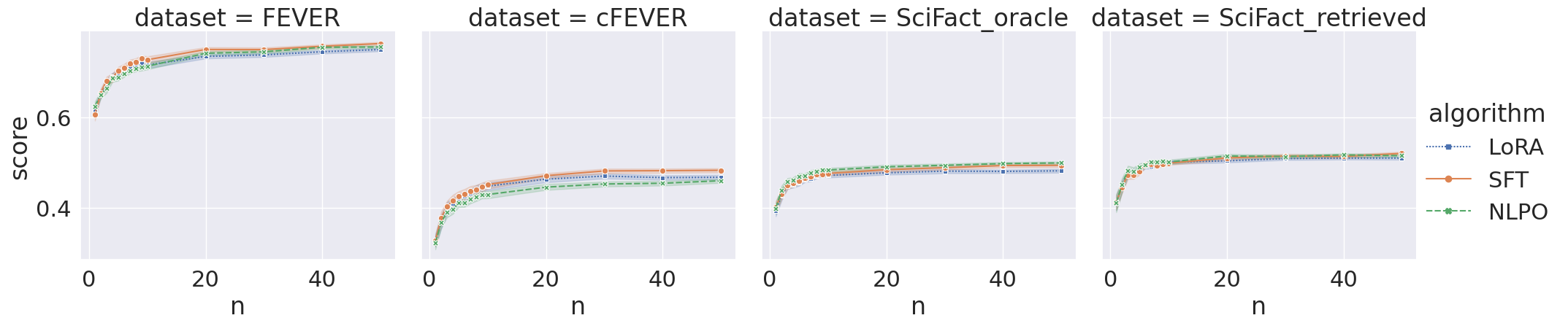}
    \caption{Comparison of MAPLE performance using different training algorithms for in-domain seq2seq training. The label ``LoRA" represents parameter-efficient training method Low-Rank Adaptation, ``SFT" indicates supervised fine-tuning and ``NLPO" refers to reinforcement learning with the NLPO policy.}
    \label{fig:ablation_algorithm}
\end{figure*}

\begin{figure*}[htb]
    \centering
    \includegraphics[scale=0.27]{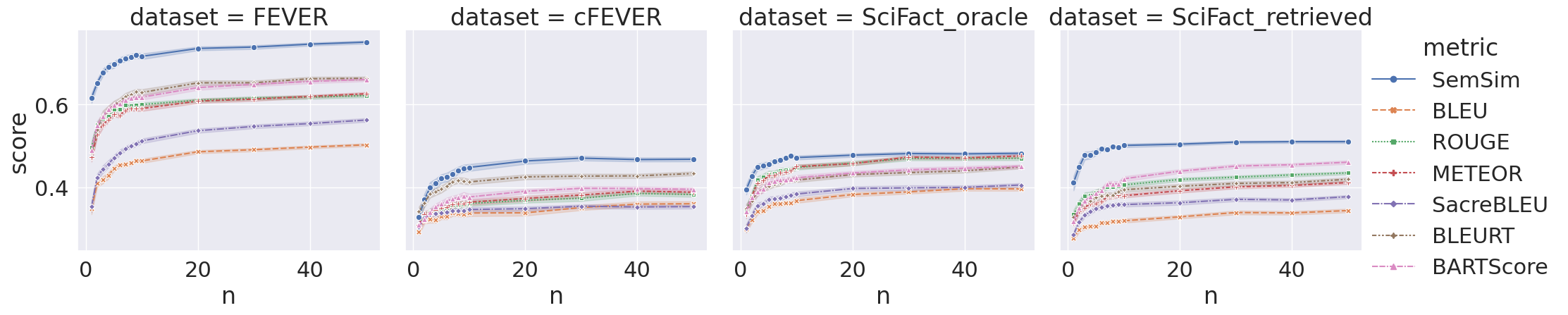}
    \caption{Comparison of MAPLE performance using the proposed `SemSim' metric and alternative metrics to measure micro pairwise language evolution. 
    }
    \label{fig:ablation_metric}
\end{figure*}
\paragraph{Training algorithms} With the growing interest in reinforcement learning (RL) and parameter-efficient training, this ablation study investigates the effects of utilizing different training algorithms. Specifically, we comprare LoRA, Supervised Fine-Tuning (SFT) and  Natural Language Policy Optimization (NLPO), an innovative RL method that offers enhanced stability and performance compared to previous policy gradient methods \cite{ramamurthy_is_2023}. As presented in Figure \ref{fig:ablation_metric}, the overall differences in performance among the algorithms are relatively marginal. SFT demonstrates best results on the FEVER and cFEVER datasets, while NLPO outperforms on the SciFact\_oracle and SciFact\_retrieved datasets. Notably, despite the largely reduced computational burden by utilizing LoRA,\footnote{For T5-small, the trainable \% with LoRA is ~0.485 (294,912/60,801,536). Please see a detailed efficiency comparison with SFT in Appendix \ref{appendix: LoRA vs SFT runtime}.} the observed performance drops are modest. Therefore, MAPLE conducts in-domain seq2seq training with LoRA.


\paragraph{Metrics} MAPLE uses our proposed `SemSim' metric to measure and analyze the pairwise language evolution. This ablation section presents the comparison with a number of established NLG metrics, including `BLEU', `ROUGE', `METEOR', `SacreBLEU', `BLEURT', and `BARTScore'.

Figure \ref{fig:ablation_metric} illustrates the performance variations of MAPLE when employing different metrics. Across all datasets, the `SemSim' metric demonstrates superior performance compared to other metrics, showcasing a significant improvement gap. This highlights the advantages of `SemSim', establishing it as the optimal choice for MAPLE. By focusing on measuring semantic similarity as a primary component, we can effectively analyze the micro pairwise evolution of language in a seq2seq learning process, which is captured by generated mutations across training epochs. In contrast, metrics based solely on lexical overlap, or utilizing an LLM that is not trained on substantial sentence pair data, may be less indicative in capturing the nuances of language evolution. The emphasis on fine-grained semantic similarity provides highly informative insights, particularly in assessing the learning difficulty of instances for seq2seq generation. As `SemSim' surpasses many established NLG metrics in this task, it shows its potential for broader applications as a general NLG evaluation metric.

\section{Analysis and Discussion}


\begin{figure*}[htb]
    \centering
    \includegraphics[scale=0.26]{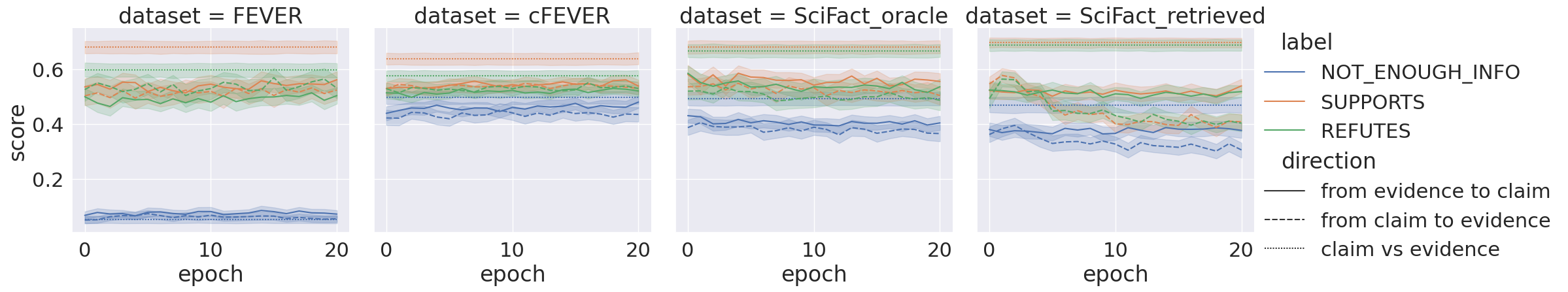}
    \caption{Example signals captured for classification, using the `SemSim' score for target-mutation pairs on the test.}
    \label{fig:interpretability}
\end{figure*}

Despite recent research on generating rationales and explanations \cite{atanasova-etal-2020-generating-fact, kotonya-toni-2020-explainable-automated, schuster-etal-2021-get}, existing approaches heavily depend on directly fine-tuning PLMs, hindering the understanding of their decision-making process. MAPLE stands out by providing tangible and traceable solutions, guided by the principle that sentence pairs with different relations present distinct challenges for seq2seq generation. Figure \ref{fig:interpretability} further supports this principle and elucidates the effectiveness of MAPLE. Overall, the `SemSim' scores for `NOT\_ENOUGH\_INFO' are significantly lower than those for `SUPPORTS' and `REFUTES', enabling easy differentiation between `NOT\_ENOUGH\_INFO' and other classes \footnote{The detailed classwise performance in Appendix \ref{appendix: MAPLE Classwise Performance within 5 shots} shows that MAPLE has the best performance on `NOT\_ENOUGH\_INFO' class.}. Furthermore, generating a piece of evidence from a claim proves to be more challenging than generating a claim from a piece of evidence. Generating claims primarily needs the removal of redundant or unnecessary content, while generating evidence requires the model to expand the existing content. Furthermore, figure \ref{fig:interpretability} shows that generating a claim is easier for `SUPPORTS' than for `REFUTES', while generating evidence is easier for `REFUTES' than for `SUPPORTS'. This pattern allows for a distinction between the two categories. With its enhanced interpretability and traceability, MAPLE aims to bolster the reliability and trustworthiness of the claim verification process.

Moreover, by comparing the difficulty among datasets based on the above information, we can gain insights into the varying challenges posed by different domains. For example, if a dataset such as FEVER consistently exhibits high `SemSim' scores and low standard deviation during in-domain seq2seq training, it suggests that the claims and evidences within that dataset are easier to match and converge upon. On the other hand, datasets such as cFEVER with lower `SemSim' scores, higher standard deviation, and longer convergence time indicate greater difficulty in aligning claims and evidences. This comparative analysis allows us to understand the relative complexities of fact-checking in different settings and further enhances the interpretability of MAPLE's performance across datasets.

Moreover, MAPLE's low demand on annotations and computing facilities enhances its efficiency and accessibility. Both step (1) in-domain seq2seq training and step (2) SemSim transformation only require unlabeled claim-evidence pairs and limited annotations are only required for step (3) logistic classifier training with few-shot labelled data. While performing steps (1) and (2) over the entire unlabeled pool may seem burdensome, such practice only takes from minutes to few hours.\footnote{Please see detailed overall runtime report in Appendix \ref{appendix: overall runtime}.} Due to MAPLE's efficiency and accessibility by design, training and deploying can be easily accomplished on Google Colab with a free account or even on a personal laptop. In real-world scenarios where the claim verification team has accumulated a substantial collection of claim-evidence pairs, which can be claims with annotated oracle evidences or claims with retrieved noisy evidences, they can initiate steps (1) and (2) and this process can be completed while the team actively acquires a small number of labeled samples. Subsequently, step (3) training a logistic classifier with the newly acquired data only takes seconds and MAPLE is ready for deployment. By designing such an efficient workflow, the application of MAPLE in real-world scenarios can bring in a decent claim verification model with minimal cost in annotation and computational resources. Overall, MAPLE holds practical value for fact-checking in real-world contexts, particularly as a tool to assist fact-checkers in combating emerging domains of misinformation.

\section{Future Directions}
With the development of MAPLE, several promising directions for future research emerge:

\paragraph{Self-supervised Extensions} Currently, MAPLE combines language transition signals with a traditional logistic classifier for classification. A further research avenue could include its development into a fully self-supervised system by integrating clustering methods.

\paragraph{NLG metric Adaptability} 
While we propose `SemSim' as an NLG metric and have demonstrated its performance advantages for MAPLE, a comprehensive evaluation of `SemSim' for broader tasks and domains would enhance the understanding. 

Most prevalent NLG evaluation metrics currently calculate similarity scores based on sentence embeddings only, including the proposed metric `SemSim' in this paper, whereas MAPLE offers nuanced insights derived from the seq2seq training dynamics. Converting MAPLE, which combines `SemSim' and T5 training, into a general NLG evaluation metric would be a promising research direction.

\paragraph{Human-in-the-loop Workflow} As previously demonstrated, MAPLE shows potential for assisting fact-checkers in real-world scenarios. Fully exploring this potential primarily involves leveraging MAPLE as a claim verification model in fact-checking organizations. Additionally, it can serve as the backbone of an active learning system, facilitating data annotation prioritization.

\section{Conclusions}
In this paper, we introduce MAPLE, a novel approach for few-shot claim verification. By leveraging language transition signals during seq2seq training convergence, MAPLE achieves SOTA performance in precisely predicting claim veracity labels with reference to associated evidences in few-shot learning scenarios. Through extensive experiments and analysis on multiple datasets, we validate its effectiveness, robustness, interpretability, efficiency and accesibility.

\section*{Limitations}
The model demonstrates quick convergence, which makes it more suitable for few-shot settings. To expand the applicability of MAPLE to higher-shot scenarios, further research and improvements are required.

\section*{Ethics Statement}
We declare that there are no conflicts of interest, ethical concerns, or potential risks associated with this work. All of the used scientifc artifacts are public open-source artifacts that are under licenses such as Apache License 2.0 and CC-BY 4.0 License and our use is consistent with their intended use. All used data does not contain any information that names or uniquely identifies individual people or offensive content and has been manually checked by the authors.

\section*{Acknowledgements}

Xia Zeng is funded by China Scholarship Council (CSC). Arkaitz Zubiaga acknowledges support from the European Union and UK Research and Innovation under Grant No. 101073351 as part of Marie Skłodowska-Curie Actions (MSCA Hybrid Intelligence to monitor, promote, and analyze transformations in good democracy practices). This research utilised Queen Mary’s Apocrita HPC facility, supported by QMUL Research-IT. \url{http://doi.org/10.5281/zenodo.438045}

\bibliography{references}
\bibliographystyle{acl_natbib}

\appendix
\label{sec:appendix}

\section{Datasets Appendix}
\label{appendix: datasets}

Table \ref{tab:datasets} shows label distributions and Table \ref{tab:examples} presents data samples for each dataset.
\begin{table*}[htbp]
    \tiny
    \centering
    \begin{tabular}{p{3.25cm}p{9.5cm}p{1cm}}
        \toprule
        \multicolumn{3}{c}{\textbf{FEVER}} \\
        \midrule
        \multicolumn{1}{c}{\textbf{Claim}} & \multicolumn{1}{c}{\textbf{Evidence}} & \multicolumn{1}{c}{\textbf{Veracity}} \\
        \midrule
        ``In 2015, among Americans, more than 50\% of adults had consumed alcoholic drink at some point.'' &
        ``For instance, in 2015, among Americans, 89\% of adults had consumed alcohol at some point, 70\% had drunk it in the last year, and 56\% in the last month.'' &
        \textit{`SUPPORTS'} \\
        \midrule
        ``Dissociative identity disorder is known only in the United States of America.'' &
        ``DID is diagnosed more frequently in North America than in the rest of the world, and is diagnosed three to nine times more often in females than in males.'' &
        \textit{`REFUTES'} \\
        \midrule
        ``Freckles induce neuromodulation.'' &
        ``Margarita Sharapova (born 15 April 1962) is a Russian novelist and short story writer whose tales often draw on her former experience as an animal trainer in a circus.'' &
        \textit{`NOT\_ ENOUGH\_ INFO'} \\
        \bottomrule
                \toprule
        \multicolumn{3}{c}{\textbf{cFEVER}} \\
        \midrule
        \multicolumn{1}{c}{\textbf{Claim}} & \multicolumn{1}{c}{\textbf{Evidence}} & \multicolumn{1}{c}{\textbf{Veracity}} \\
        \midrule
        ``Coral atolls grow as sea levels rise.'' &
        ``Gradual sea-level rise also allows for coral polyp activity to raise the atolls with the sea level.'' &
        \textit{`SUPPORTS'} \\
        \midrule
        ``There’s no trend in hurricane-related flooding in the U.S.'' &
        ``Widespread heavy rainfall contributed to significant inland flooding from Louisiana into Arkansas.'' &
        \textit{`REFUTES'} \\
        \midrule
        ``The warming is not nearly as great as the climate change computer models have predicted.'' &
        ``The model predicted <0.2 °C warming for upper air at 700 mb and 500 mb.'' &
        \textit{`NOT\_ ENOUGH\_ INFO'} \\
        \bottomrule
        \toprule
        \multicolumn{3}{c}{\textbf{SCIFACT\_oracle}} \\
        \midrule
        \multicolumn{1}{c}{\textbf{Claim}} & \multicolumn{1}{c}{\textbf{Evidence}} & \multicolumn{1}{c}{\textbf{Veracity}} \\
        \midrule
        ``Macropinocytosis contributes to a cell's supply of amino acids via the intracellular uptake of protein.'' &
        ``Here, we demonstrate that protein macropinocytosis can also serve as an essential amino acid source.'' &
        \textit{`SUPPORTS'} \\
        \midrule
        ``Gene expression does not vary appreciably across genetically identical cells.'' &
        ``Genetically identical cells sharing an environment can display markedly different phenotypes.'' &
        \textit{`REFUTES'} \\
        \midrule
        ``Fz/PCP-dependent Pk localizes to the anterior membrane of notochord cells during zebrafish neuralation.'' &
        ``These results reveal a function for PCP signalling in coupling cell division and morphogenesis at neurulation and indicate a previously unrecognized mechanism that might underlie NTDs.'' &
        \textit{`NOT\_ ENOUGH\_ INFO'} \\
        \bottomrule
        \toprule
        \multicolumn{3}{c}{\textbf{SCIFACT\_retrieved}} \\
        \midrule
        \multicolumn{1}{c}{\textbf{Claim}} & \multicolumn{1}{c}{\textbf{Evidence}} & \multicolumn{1}{c}{\textbf{Veracity}} \\
        \midrule
        ``Neutrophil extracellular trap (NET) antigens may contain the targeted autoantigens PR3 and MPO.'' &
        ``Netting neutrophils in autoimmune small-vessel vasculitis Small-vessel vasculitis (SVV) is a chronic autoinflammatory condition linked to antineutrophil cytoplasm autoantibodies (ANCAs). Here we show that chromatin fibers, so-called neutrophil extracellular traps (NETs), are released by ANCA-stimulated neutrophils and contain the targeted autoantigens proteinase-3 (PR3) and myeloperoxidase (MPO). Deposition of NETs in inflamed kidneys and circulating MPO-DNA complexes suggest that NET formation triggers vasculitis and promotes the autoimmune response against neutrophil components in individuals with SVV.'' &
        \textit{`SUPPORTS'} \\
        \midrule
        ``Cytochrome c is transferred from cytosol to the mitochondrial intermembrane space during apoptosis.'' &
        ``At the gates of death. Apoptosis that proceeds via the mitochondrial pathway involves mitochondrial outer membrane permeabilization (MOMP), responsible for the release of cytochrome c and other proteins of the mitochondrial intermembrane space. This essential step is controlled and mediated by proteins of the Bcl-2 family. The proapoptotic proteins Bax and Bak are required for MOMP, while the antiapoptotic Bcl-2 proteins, including Bcl-2, Bcl-xL, Mcl-1, and others, prevent MOMP. Different proapoptotic BH3-only proteins act to interfere with the function of the antiapoptotic Bcl-2 members and\/or activate Bax and Bak. Here, we discuss an emerging view, proposed by Certo et al. in this issue of Cancer Cell, on how these interactions result in MOMP and apoptosis.'' &
        \textit{`REFUTES'} \\
        \midrule
        ``Incidence of heart failure increased by 10\% in women since 1979.'' &
        ``Clinical epidemiology of heart failure. The aim of this paper is to review the clinical epidemiology of heart failure. The last paper comprehensively addressing the epidemiology of heart failure in Heart appeared in 2000. Despite an increase in manuscripts describing epidemiological aspects of heart failure since the 1990s, additional information is still needed, as indicated by various editorials.'' &
        \textit{`NOT\_ ENOUGH\_ INFO'} \\
        \bottomrule

    \end{tabular}
    \caption{Data samples for each dataset.}
    \label{tab:examples}
\end{table*}

\begin{table*}[htbp]
  \centering
  \caption{Unlabelled pool label distribution for each dataset.}
  \label{tab:datasets}
  \begin{tabular}{lcccc}
    \hline
    & \textbf{FEVER} & \textbf{cFEVER} & \textbf{SciFact\_oracle} & \textbf{SciFact\_retrieved} \\
    \hline
    \textbf{`SUPPORTS'} & 3099 & 1789 & 356 & 266 \\
    \hline
    \textbf{`REFUTES'} & 3069 & 652 & 115 & 61 \\
    \hline
    \textbf{`NOT\_ENOUGH\_INFO'} & 3183 & 4778 & 294 & 2530 \\
    \hline
    \textbf{Total unlabelled pairs} & 9351 & 7219 & 765 & 2857 \\
    \hline
  \end{tabular}
\end{table*}

\section{Performance Appendix}
\subsection{Detailed performance comparison across methods}
\label{appendix: detailed performance comparison}

Here we present a detailed numeric performance comparison of the methods discussed, as well as alternative model checkpoints for PET\footnote{We report all six model checkpoints used in Active PETs.} and LLaMA 2\footnote{We report all three models that have chat capabilities.}.\footnote{When the same prompt we deigned for 7b model is used on 13b and 70b models, the model performance is significantly lower and even fails to yield responses in many cases and vise versa. Hence, the results for 13b and 70b models in this section are generated with a prompt that is slightly different from the one we used for 7b model. The prompt we used here is ``Please perform the task of claim verification. Given a claim and a piece of evidence, your goal is to classify them into one of the following classes: `SUPPORTS', `REFUTES' and `NOT\_ENOUGH\_INFO'. Here are a few examples: Claim: `{train\_claim\_i}' Evidence: `{train\_evidences\_i}' `{train\_labels\_i}'.''. The post-process remains the same.} Tables \ref{tab:FEVER detailed}, \ref{tab:cFEVER detailed}, \ref{tab:SciFactoracle detailed} and \ref{tab:SciFactretrieved detailed} report on FEVER, cFEVER, SciFact\_oracle and SciFact\_retrieved dataset configurations respectively.

\begin{table*}[htbp]
\small
\centering
\begin{tabular}{llrrrr}
\toprule
  FEVER &                                & \multicolumn{2}{l}{F1} & \multicolumn{2}{l}{Accuracy} \\
n-shot & method                        &    mean &     std &    mean &     std \\
\midrule
1 & Llama-2-7b-chat-hf &  0.3776 &  0.0438 &  0.4771 &  0.0439 \\
  & Llama-2-13b-chat-hf &  0.4351 &  0.0613 &  0.5034 &  0.0506 \\
  & Llama-2-70b-chat-hf &  0.2617 &  0.0427 &  0.3800 &  0.0258 \\
  & MAPLE &  0.6155 &  0.0645 &  0.6459 &  0.0506 \\
  & PET\_microsoft/deberta-base-mnli &  0.3394 &  0.0351 &  0.3582 &  0.0293 \\
  & PET\_microsoft/deberta-large-mnli &  0.4978 &  0.1011 &  0.5193 &  0.0877 \\
  & PET\_roberta-large-mnli &  0.2158 &  0.0516 &  0.2408 &  0.0670 \\
  & PET\_textattack/bert-base-uncased-MNLI &  0.3731 &  0.0456 &  0.4089 &  0.0278 \\
  & PET\_textattack/roberta-base-MNLI &  0.2190 &  0.0409 &  0.3139 &  0.0383 \\
  & PET\_yoshitomo-matsubara/bert-large-uncased-mnli &  0.4214 &  0.0480 &  0.4509 &  0.0429 \\
  & SEED\_bert-base-nli-mean-tokens &  0.2724 &  0.0689 &  0.3748 &  0.0494 \\
2 & Llama-2-7b-chat-hf &  0.3827 &  0.0301 &  0.4796 &  0.0314 \\
  & Llama-2-13b-chat-hf &  0.3929 &  0.0504 &  0.4719 &  0.0393 \\
  & Llama-2-70b-chat-hf &  0.2745 &  0.0402 &  0.3883 &  0.0256 \\
  & MAPLE &  0.6514 &  0.0460 &  0.6724 &  0.0379 \\
  & PET\_microsoft/deberta-base-mnli &  0.3773 &  0.0354 &  0.3870 &  0.0374 \\
  & PET\_microsoft/deberta-large-mnli &  0.5897 &  0.0917 &  0.6023 &  0.0843 \\
  & PET\_roberta-large-mnli &  0.2308 &  0.0463 &  0.2526 &  0.0617 \\
  & PET\_textattack/bert-base-uncased-MNLI &  0.4151 &  0.0372 &  0.4338 &  0.0261 \\
  & PET\_textattack/roberta-base-MNLI &  0.2661 &  0.0408 &  0.3349 &  0.0340 \\
  & PET\_yoshitomo-matsubara/bert-large-uncased-mnli &  0.4689 &  0.0490 &  0.4904 &  0.0448 \\
  & SEED\_bert-base-nli-mean-tokens &  0.3935 &  0.0822 &  0.4455 &  0.0667 \\
3 & Llama-2-7b-chat-hf &  0.3760 &  0.0321 &  0.4702 &  0.0312 \\
  & Llama-2-13b-chat-hf &  0.3815 &  0.0371 &  0.4606 &  0.0299 \\
  & Llama-2-70b-chat-hf &  0.2792 &  0.0379 &  0.3930 &  0.0246 \\
  & MAPLE &  0.6768 &  0.0448 &  0.6911 &  0.0400 \\
  & PET\_microsoft/deberta-base-mnli &  0.3977 &  0.0327 &  0.4069 &  0.0315 \\
  & PET\_microsoft/deberta-large-mnli &  0.6586 &  0.0768 &  0.6649 &  0.0733 \\
  & PET\_roberta-large-mnli &  0.2551 &  0.0406 &  0.2682 &  0.0513 \\
  & PET\_textattack/bert-base-uncased-MNLI &  0.4429 &  0.0267 &  0.4524 &  0.0213 \\
  & PET\_textattack/roberta-base-MNLI &  0.2810 &  0.0361 &  0.3389 &  0.0330 \\
  & PET\_yoshitomo-matsubara/bert-large-uncased-mnli &  0.4999 &  0.0401 &  0.5186 &  0.0367 \\
  & SEED\_bert-base-nli-mean-tokens &  0.4843 &  0.0714 &  0.5118 &  0.0615 \\
4 & Llama-2-7b-chat-hf &  0.3621 &  0.0473 &  0.4562 &  0.0408 \\
  & Llama-2-13b-chat-hf &  0.3790 &  0.0425 &  0.4598 &  0.0343 \\
  & Llama-2-70b-chat-hf &  0.2874 &  0.0382 &  0.3988 &  0.0248 \\
  & MAPLE &  0.6909 &  0.0399 &  0.7019 &  0.0368 \\
  & PET\_microsoft/deberta-base-mnli &  0.4142 &  0.0292 &  0.4203 &  0.0293 \\
  & PET\_microsoft/deberta-large-mnli &  0.6893 &  0.0628 &  0.6943 &  0.0603 \\
  & PET\_roberta-large-mnli &  0.2786 &  0.0405 &  0.2993 &  0.0517 \\
  & PET\_textattack/bert-base-uncased-MNLI &  0.4623 &  0.0211 &  0.4667 &  0.0186 \\
  & PET\_textattack/roberta-base-MNLI &  0.3000 &  0.0353 &  0.3445 &  0.0326 \\
  & PET\_yoshitomo-matsubara/bert-large-uncased-mnli &  0.5191 &  0.0364 &  0.5318 &  0.0326 \\
  & SEED\_bert-base-nli-mean-tokens &  0.5331 &  0.0619 &  0.5495 &  0.0568 \\
5 & Llama-2-7b-chat-hf &  0.3613 &  0.0468 &  0.4472 &  0.0367 \\
  & Llama-2-13b-chat-hf &  0.3781 &  0.0320 &  0.4592 &  0.0275 \\
  & Llama-2-70b-chat-hf &  0.2997 &  0.0371 &  0.4074 &  0.0247 \\
  & MAPLE &  0.6964 &  0.0403 &  0.7058 &  0.0368 \\
  & PET\_microsoft/deberta-base-mnli &  0.4266 &  0.0270 &  0.4320 &  0.0274 \\
  & PET\_microsoft/deberta-large-mnli &  0.7191 &  0.0584 &  0.7237 &  0.0564 \\
  & PET\_roberta-large-mnli &  0.2941 &  0.0396 &  0.3188 &  0.0443 \\
  & PET\_textattack/bert-base-uncased-MNLI &  0.4699 &  0.0173 &  0.4731 &  0.0153 \\
  & PET\_textattack/roberta-base-MNLI &  0.3064 &  0.0293 &  0.3456 &  0.0293 \\
  & PET\_yoshitomo-matsubara/bert-large-uncased-mnli &  0.5267 &  0.0358 &  0.5410 &  0.0318 \\
  & SEED\_bert-base-nli-mean-tokens &  0.5714 &  0.0556 &  0.5821 &  0.0538 \\
\bottomrule
\end{tabular}
\caption{Detailed performance on FEVER. The reported results are mean and standard deviation for F1 and accuracy scores on 100 runs.}
\label{tab:FEVER detailed}
\end{table*}

\begin{table*}[htbp]
\small
\centering
\begin{tabular}{llrrrr}
\toprule
  cFEVER &                                & \multicolumn{2}{l}{F1} & \multicolumn{2}{l}{Accuracy} \\
n-shot & method                        &    mean &     std &    mean &     std \\
\midrule
1 & Llama-2-7b-chat-hf &  0.3798 &  0.0346 &  0.4184 &  0.0226 \\
  & Llama-2-13b-chat-hf &  0.4769 &  0.0380 &  0.4831 &  0.0345 \\
  & Llama-2-70b-chat-hf &  0.2793 &  0.0439 &  0.3620 &  0.0263 \\
  & MAPLE &  0.3276 &  0.0717 &  0.3622 &  0.0696 \\
  & PET\_microsoft/deberta-base-mnli &  0.2401 &  0.0209 &  0.3072 &  0.0221 \\
  & PET\_microsoft/deberta-large-mnli &  0.3519 &  0.0672 &  0.3795 &  0.0657 \\
  & PET\_roberta-large-mnli &  0.2828 &  0.0594 &  0.3078 &  0.0555 \\
  & PET\_textattack/bert-base-uncased-MNLI &  0.2721 &  0.0198 &  0.3151 &  0.0159 \\
  & PET\_textattack/roberta-base-MNLI &  0.1850 &  0.0103 &  0.3175 &  0.0166 \\
  & PET\_yoshitomo-matsubara/bert-large-uncased-mnli &  0.3519 &  0.0382 &  0.3782 &  0.0302 \\
  & SEED\_bert-base-nli-mean-tokens &  0.2834 &  0.0621 &  0.3640 &  0.0464 \\
2 & Llama-2-7b-chat-hf &  0.3541 &  0.0228 &  0.4067 &  0.0180 \\
  & Llama-2-13b-chat-hf &  0.3745 &  0.0602 &  0.4007 &  0.0390 \\
  & Llama-2-70b-chat-hf &  0.2481 &  0.0363 &  0.3389 &  0.0209 \\
  & MAPLE &  0.3700 &  0.0788 &  0.3899 &  0.0748 \\
  & PET\_microsoft/deberta-base-mnli &  0.2574 &  0.0175 &  0.3069 &  0.0215 \\
  & PET\_microsoft/deberta-large-mnli &  0.3958 &  0.0633 &  0.4148 &  0.0581 \\
  & PET\_roberta-large-mnli &  0.3147 &  0.0615 &  0.3329 &  0.0597 \\
  & PET\_textattack/bert-base-uncased-MNLI &  0.2898 &  0.0172 &  0.3129 &  0.0162 \\
  & PET\_textattack/roberta-base-MNLI &  0.1962 &  0.0159 &  0.3199 &  0.0200 \\
  & PET\_yoshitomo-matsubara/bert-large-uncased-mnli &  0.3621 &  0.0364 &  0.3846 &  0.0268 \\
  & SEED\_bert-base-nli-mean-tokens &  0.3574 &  0.0621 &  0.4020 &  0.0538 \\
3 & Llama-2-7b-chat-hf &  0.3638 &  0.0287 &  0.4041 &  0.0188 \\
  & Llama-2-13b-chat-hf &  0.3866 &  0.0534 &  0.4091 &  0.0359 \\
  & Llama-2-70b-chat-hf &  0.2515 &  0.0333 &  0.3448 &  0.0153 \\
  & MAPLE &  0.3993 &  0.0678 &  0.4112 &  0.0643 \\
  & PET\_microsoft/deberta-base-mnli &  0.2665 &  0.0179 &  0.3059 &  0.0190 \\
  & PET\_microsoft/deberta-large-mnli &  0.4081 &  0.0601 &  0.4215 &  0.0603 \\
  & PET\_roberta-large-mnli &  0.3278 &  0.0565 &  0.3448 &  0.0549 \\
  & PET\_textattack/bert-base-uncased-MNLI &  0.2965 &  0.0141 &  0.3107 &  0.0151 \\
  & PET\_textattack/roberta-base-MNLI &  0.2046 &  0.0195 &  0.3196 &  0.0230 \\
  & PET\_yoshitomo-matsubara/bert-large-uncased-mnli &  0.3675 &  0.0374 &  0.3943 &  0.0242 \\
  & SEED\_bert-base-nli-mean-tokens &  0.3857 &  0.0550 &  0.4180 &  0.0559 \\
4 & Llama-2-7b-chat-hf &  0.3662 &  0.0243 &  0.4001 &  0.0157 \\
  & Llama-2-13b-chat-hf &  0.4158 &  0.0466 &  0.4284 &  0.0388 \\
  & Llama-2-70b-chat-hf &  0.2631 &  0.0337 &  0.3514 &  0.0169 \\
  & MAPLE &  0.4089 &  0.0677 &  0.4181 &  0.0648 \\
  & PET\_microsoft/deberta-base-mnli &  0.2750 &  0.0202 &  0.3105 &  0.0198 \\
  & PET\_microsoft/deberta-large-mnli &  0.4324 &  0.0424 &  0.4456 &  0.0420 \\
  & PET\_roberta-large-mnli &  0.3504 &  0.0533 &  0.3652 &  0.0487 \\
  & PET\_textattack/bert-base-uncased-MNLI &  0.3033 &  0.0143 &  0.3141 &  0.0139 \\
  & PET\_textattack/roberta-base-MNLI &  0.2109 &  0.0196 &  0.3221 &  0.0209 \\
  & PET\_yoshitomo-matsubara/bert-large-uncased-mnli &  0.3710 &  0.0338 &  0.3972 &  0.0218 \\
  & SEED\_bert-base-nli-mean-tokens &  0.4069 &  0.0477 &  0.4344 &  0.0467 \\
5 & Llama-2-7b-chat-hf &  0.3709 &  0.0271 &  0.3932 &  0.0191 \\
  & Llama-2-13b-chat-hf &  0.4473 &  0.0417 &  0.4540 &  0.0367 \\
  & Llama-2-70b-chat-hf &  0.2752 &  0.0375 &  0.3575 &  0.0182 \\
  & MAPLE &  0.4208 &  0.0548 &  0.4299 &  0.0520 \\
  & PET\_microsoft/deberta-base-mnli &  0.2838 &  0.0198 &  0.3148 &  0.0215 \\
  & PET\_microsoft/deberta-large-mnli &  0.4488 &  0.0443 &  0.4606 &  0.0431 \\
  & PET\_roberta-large-mnli &  0.3587 &  0.0497 &  0.3751 &  0.0424 \\
  & PET\_textattack/bert-base-uncased-MNLI &  0.3049 &  0.0132 &  0.3129 &  0.0127 \\
  & PET\_textattack/roberta-base-MNLI &  0.2121 &  0.0189 &  0.3200 &  0.0208 \\
  & PET\_yoshitomo-matsubara/bert-large-uncased-mnli &  0.3719 &  0.0311 &  0.4001 &  0.0200 \\
  & SEED\_bert-base-nli-mean-tokens &  0.4164 &  0.0380 &  0.4409 &  0.0371 \\
\bottomrule
\end{tabular}
\caption{Detailed performance on cFEVER. The reported results are mean and standard deviation for F1 and accuracy scores on 100 runs.}
\label{tab:cFEVER detailed}
\end{table*}

\begin{table*}[htbp]
\small
\centering
\begin{tabular}{llrrrr}
\toprule
  SciFact\_oracle &                                & \multicolumn{2}{l}{F1} & \multicolumn{2}{l}{Accuracy} \\
n-shot & method                        &    mean &     std &    mean &     std \\
\midrule
\midrule
1 & Llama-2-7b-chat-hf &  0.3746 &  0.0306 &  0.4549 &  0.0295 \\
  & Llama-2-13b-chat-hf &  0.3722 &  0.0481 &  0.4359 &  0.0375 \\
  & Llama-2-70b-chat-hf &  0.2502 &  0.0417 &  0.3706 &  0.0233 \\
  & MAPLE &  0.3938 &  0.0658 &  0.4333 &  0.0604 \\
  & PET\_microsoft/deberta-base-mnli &  0.2459 &  0.0244 &  0.3112 &  0.0121 \\
  & PET\_microsoft/deberta-large-mnli &  0.4467 &  0.0833 &  0.4699 &  0.0735 \\
  & PET\_roberta-large-mnli &  0.2514 &  0.0537 &  0.2747 &  0.0569 \\
  & PET\_textattack/bert-base-uncased-MNLI &  0.3696 &  0.0435 &  0.4059 &  0.0314 \\
  & PET\_textattack/roberta-base-MNLI &  0.2352 &  0.0273 &  0.3338 &  0.0301 \\
  & PET\_yoshitomo-matsubara/bert-large-uncased-mnli &  0.3078 &  0.0255 &  0.3312 &  0.0257 \\
  & SEED\_bert-base-nli-mean-tokens &  0.2996 &  0.0634 &  0.3757 &  0.0489 \\
2 & Llama-2-7b-chat-hf &  0.3812 &  0.0233 &  0.4678 &  0.0237 \\
  & Llama-2-13b-chat-hf &  0.3489 &  0.0382 &  0.4180 &  0.0313 \\
  & Llama-2-70b-chat-hf &  0.2614 &  0.0329 &  0.3698 &  0.0176 \\
  & MAPLE &  0.4263 &  0.0571 &  0.4493 &  0.0575 \\
  & PET\_microsoft/deberta-base-mnli &  0.2686 &  0.0170 &  0.3152 &  0.0120 \\
  & PET\_microsoft/deberta-large-mnli &  0.5099 &  0.0772 &  0.5265 &  0.0673 \\
  & PET\_roberta-large-mnli &  0.2824 &  0.0503 &  0.3014 &  0.0569 \\
  & PET\_textattack/bert-base-uncased-MNLI &  0.3973 &  0.0337 &  0.4218 &  0.0266 \\
  & PET\_textattack/roberta-base-MNLI &  0.2534 &  0.0280 &  0.3378 &  0.0304 \\
  & PET\_yoshitomo-matsubara/bert-large-uncased-mnli &  0.3068 &  0.0279 &  0.3401 &  0.0196 \\
  & SEED\_bert-base-nli-mean-tokens &  0.3552 &  0.0648 &  0.3937 &  0.0600 \\
3 & Llama-2-7b-chat-hf &  0.3998 &  0.0377 &  0.4662 &  0.0281 \\
  & Llama-2-13b-chat-hf &  0.3475 &  0.0395 &  0.4112 &  0.0315 \\
  & Llama-2-70b-chat-hf &  0.2739 &  0.0377 &  0.3753 &  0.0227 \\
  & MAPLE &  0.4487 &  0.0402 &  0.4655 &  0.0384 \\
  & PET\_microsoft/deberta-base-mnli &  0.2841 &  0.0163 &  0.3237 &  0.0120 \\
  & PET\_microsoft/deberta-large-mnli &  0.5508 &  0.0722 &  0.5639 &  0.0637 \\
  & PET\_roberta-large-mnli &  0.2936 &  0.0448 &  0.3159 &  0.0516 \\
  & PET\_textattack/bert-base-uncased-MNLI &  0.4153 &  0.0253 &  0.4312 &  0.0197 \\
  & PET\_textattack/roberta-base-MNLI &  0.2633 &  0.0256 &  0.3372 &  0.0276 \\
  & PET\_yoshitomo-matsubara/bert-large-uncased-mnli &  0.3047 &  0.0258 &  0.3427 &  0.0181 \\
  & SEED\_bert-base-nli-mean-tokens &  0.4007 &  0.0593 &  0.4290 &  0.0593 \\
4 & Llama-2-7b-chat-hf &  0.4002 &  0.0420 &  0.4542 &  0.0312 \\
  & Llama-2-13b-chat-hf &  0.3558 &  0.0365 &  0.4165 &  0.0306 \\
  & Llama-2-70b-chat-hf &  0.2939 &  0.0454 &  0.3888 &  0.0277 \\
  & MAPLE &  0.4520 &  0.0426 &  0.4661 &  0.0405 \\
  & PET\_microsoft/deberta-base-mnli &  0.2932 &  0.0180 &  0.3265 &  0.0132 \\
  & PET\_microsoft/deberta-large-mnli &  0.5698 &  0.0738 &  0.5781 &  0.0677 \\
  & PET\_roberta-large-mnli &  0.2988 &  0.0540 &  0.3173 &  0.0585 \\
  & PET\_textattack/bert-base-uncased-MNLI &  0.4197 &  0.0220 &  0.4361 &  0.0157 \\
  & PET\_textattack/roberta-base-MNLI &  0.2743 &  0.0263 &  0.3416 &  0.0287 \\
  & PET\_yoshitomo-matsubara/bert-large-uncased-mnli &  0.3054 &  0.0269 &  0.3461 &  0.0187 \\
  & SEED\_bert-base-nli-mean-tokens &  0.4289 &  0.0519 &  0.4499 &  0.0503 \\
5 & Llama-2-7b-chat-hf &  0.3998 &  0.0463 &  0.4487 &  0.0328 \\
  & Llama-2-13b-chat-hf &  0.3611 &  0.0348 &  0.4231 &  0.0308 \\
  & Llama-2-70b-chat-hf &  0.2840 &  0.0709 &  0.3873 &  0.0370 \\
  & MAPLE &  0.4554 &  0.0356 &  0.4675 &  0.0356 \\
  & PET\_microsoft/deberta-base-mnli &  0.3005 &  0.0172 &  0.3312 &  0.0139 \\
  & PET\_microsoft/deberta-large-mnli &  0.5964 &  0.0706 &  0.6045 &  0.0641 \\
  & PET\_roberta-large-mnli &  0.3087 &  0.0507 &  0.3281 &  0.0558 \\
  & PET\_textattack/bert-base-uncased-MNLI &  0.4252 &  0.0233 &  0.4413 &  0.0147 \\
  & PET\_textattack/roberta-base-MNLI &  0.2780 &  0.0222 &  0.3420 &  0.0249 \\
  & PET\_yoshitomo-matsubara/bert-large-uncased-mnli &  0.3072 &  0.0274 &  0.3496 &  0.0166 \\
  & SEED\_bert-base-nli-mean-tokens &  0.4463 &  0.0478 &  0.4645 &  0.0465 \\
\bottomrule
\end{tabular}
\caption{Detailed performance on SciFact\_oracle. The reported results are mean and standard deviation for F1 and accuracy scores on 100 runs.}
\label{tab:SciFactoracle detailed}
\end{table*}

\begin{table*}[htbp]
\small
\centering
\begin{tabular}{llrrrr}
\toprule
  SciFact\_retrieved &                                & \multicolumn{2}{l}{F1} & \multicolumn{2}{l}{Accuracy} \\
n-shot & method                        &    mean &     std &    mean &     std \\
\midrule
1 & Llama-2-7b-chat-hf &  0.3207 &  0.0299 &  0.3943 &  0.0243 \\
  & Llama-2-13b-chat-hf &  0.3757 &  0.0380 &  0.4265 &  0.0231 \\
  & Llama-2-70b-chat-hf &  0.3454 &  0.0598 &  0.4035 &  0.0338 \\
  & MAPLE &  0.4108 &  0.0878 &  0.4412 &  0.0831 \\
  & PET\_microsoft/deberta-base-mnli &  0.2927 &  0.0341 &  0.3134 &  0.0302 \\
  & PET\_microsoft/deberta-large-mnli &  0.3332 &  0.0525 &  0.3609 &  0.0450 \\
  & PET\_roberta-large-mnli &  0.2448 &  0.0308 &  0.2830 &  0.0298 \\
  & PET\_textattack/bert-base-uncased-MNLI &  0.3431 &  0.0263 &  0.3661 &  0.0180 \\
  & PET\_textattack/roberta-base-MNLI &  0.2598 &  0.0317 &  0.3491 &  0.0238 \\
  & PET\_yoshitomo-matsubara/bert-large-uncased-mnli &  0.3162 &  0.0352 &  0.3477 &  0.0215 \\
  & SEED\_bert-base-nli-mean-tokens &  0.2708 &  0.0470 &  0.3479 &  0.0288 \\
2 & Llama-2-7b-chat-hf &  0.2914 &  0.0528 &  0.3586 &  0.0350 \\
  & Llama-2-13b-chat-hf &  0.3278 &  0.0524 &  0.3925 &  0.0266 \\
  & Llama-2-70b-chat-hf &  0.1682 &  0.0105 &  0.3338 &  0.0038 \\
  & MAPLE &  0.4484 &  0.0699 &  0.4654 &  0.0675 \\
  & PET\_microsoft/deberta-base-mnli &  0.2988 &  0.0315 &  0.3147 &  0.0281 \\
  & PET\_microsoft/deberta-large-mnli &  0.3601 &  0.0524 &  0.3834 &  0.0434 \\
  & PET\_roberta-large-mnli &  0.2576 &  0.0300 &  0.2891 &  0.0281 \\
  & PET\_textattack/bert-base-uncased-MNLI &  0.3514 &  0.0201 &  0.3633 &  0.0179 \\
  & PET\_textattack/roberta-base-MNLI &  0.2944 &  0.0289 &  0.3549 &  0.0267 \\
  & PET\_yoshitomo-matsubara/bert-large-uncased-mnli &  0.3156 &  0.0333 &  0.3571 &  0.0199 \\
  & SEED\_bert-base-nli-mean-tokens &  0.3233 &  0.0463 &  0.3623 &  0.0439 \\
3 & Llama-2-7b-chat-hf &  0.1775 &  0.0363 &  0.3329 &  0.0056 \\
  & Llama-2-13b-chat-hf &  0.1788 &  0.0371 &  0.3359 &  0.0104 \\
  & Llama-2-70b-chat-hf &  0.1667 &  0.0000 &  0.3333 &  0.0000 \\
  & MAPLE &  0.4768 &  0.0511 &  0.4909 &  0.0464 \\
  & PET\_microsoft/deberta-base-mnli &  0.2963 &  0.0308 &  0.3085 &  0.0249 \\
  & PET\_microsoft/deberta-large-mnli &  0.3599 &  0.0518 &  0.3880 &  0.0419 \\
  & PET\_roberta-large-mnli &  0.2557 &  0.0266 &  0.2853 &  0.0243 \\
  & PET\_textattack/bert-base-uncased-MNLI &  0.3490 &  0.0212 &  0.3604 &  0.0179 \\
  & PET\_textattack/roberta-base-MNLI &  0.3135 &  0.0251 &  0.3559 &  0.0250 \\
  & PET\_yoshitomo-matsubara/bert-large-uncased-mnli &  0.3102 &  0.0281 &  0.3580 &  0.0171 \\
  & SEED\_bert-base-nli-mean-tokens &  0.3530 &  0.0382 &  0.3795 &  0.0367 \\
4 & Llama-2-7b-chat-hf &  0.1667 &  0.0000 &  0.3333 &  0.0000 \\
  & Llama-2-13b-chat-hf &  0.1667 &  0.0000 &  0.3333 &  0.0000 \\
  & Llama-2-70b-chat-hf &  0.1667 &  0.0000 &  0.3333 &  0.0000 \\
  & MAPLE &  0.4777 &  0.0449 &  0.4884 &  0.0429 \\
  & PET\_microsoft/deberta-base-mnli &  0.3038 &  0.0278 &  0.3129 &  0.0252 \\
  & PET\_microsoft/deberta-large-mnli &  0.3827 &  0.0494 &  0.4026 &  0.0453 \\
  & PET\_roberta-large-mnli &  0.2616 &  0.0236 &  0.2862 &  0.0224 \\
  & PET\_textattack/bert-base-uncased-MNLI &  0.3467 &  0.0240 &  0.3611 &  0.0195 \\
  & PET\_textattack/roberta-base-MNLI &  0.3289 &  0.0284 &  0.3611 &  0.0245 \\
  & PET\_yoshitomo-matsubara/bert-large-uncased-mnli &  0.3083 &  0.0253 &  0.3582 &  0.0173 \\
  & SEED\_bert-base-nli-mean-tokens &  0.3581 &  0.0383 &  0.3820 &  0.0369 \\
5 & Llama-2-7b-chat-hf &  0.1667 &  0.0000 &  0.3333 &  0.0000 \\
  & Llama-2-13b-chat-hf &  0.1667 &  0.0000 &  0.3333 &  0.0000 \\
  & Llama-2-70b-chat-hf &  0.1667 &  0.0000 &  0.3333 &  0.0000 \\
  & MAPLE &  0.4846 &  0.0351 &  0.4941 &  0.0331 \\
  & PET\_microsoft/deberta-base-mnli &  0.3054 &  0.0261 &  0.3163 &  0.0240 \\
  & PET\_microsoft/deberta-large-mnli &  0.3825 &  0.0504 &  0.4043 &  0.0435 \\
  & PET\_roberta-large-mnli &  0.2575 &  0.0274 &  0.2915 &  0.0225 \\
  & PET\_textattack/bert-base-uncased-MNLI &  0.3467 &  0.0242 &  0.3624 &  0.0197 \\
  & PET\_textattack/roberta-base-MNLI &  0.3348 &  0.0252 &  0.3600 &  0.0226 \\
  & PET\_yoshitomo-matsubara/bert-large-uncased-mnli &  0.3066 &  0.0289 &  0.3638 &  0.0165 \\
  & SEED\_bert-base-nli-mean-tokens &  0.3726 &  0.0361 &  0.3903 &  0.0367 \\
\bottomrule
\end{tabular}
\caption{Detailed performance on SciFact\_retrieved. The reported results are mean and standard deviation for F1 and accuracy scores on 100 runs.}
\label{tab:SciFactretrieved detailed}
\end{table*}

\subsection{MAPLE Classwise Performance within 5 shots}
\label{appendix: MAPLE Classwise Performance within 5 shots}

Table \ref{tab:classwise} presents MAPLE's classwise performance. In general, MAPLE is most capable of distinguishing NOT\_ENOUGH\_INFO samples from the others and the least capable when dealing with REFUTES samples.

\begin{table*}[htbp]
    \small
    \centering
    \begin{tabular}{lcccccc}
    \toprule
    \multicolumn{7}{c}{\textbf{FEVER}} \\
    \midrule
    n-shot & \multicolumn{2}{c}{F1(SUPPORTS)} & \multicolumn{2}{c}{F1(NOT\_ENOUGH\_INFO)} & \multicolumn{2}{c}{F1(REFUTES)} \\
     &        mean &    std &               mean &    std &       mean &    std \\
    \midrule
1  &      0.4737 & 0.1665 &             0.9177 & 0.1010 &     0.4550 & 0.1557 \\
2  &      0.5144 & 0.1167 &             0.9442 & 0.0270 &     0.4955 & 0.1330 \\
3  &      0.5593 & 0.1077 &             0.9531 & 0.0193 &     0.5181 & 0.0972 \\
4  &      0.5762 & 0.0938 &             0.9550 & 0.0186 &     0.5416 & 0.0807 \\
5  &      0.5821 & 0.0891 &             0.9584 & 0.0157 &     0.5487 & 0.0805 \\
    \bottomrule
        \toprule
    \multicolumn{7}{c}{\textbf{cFEVER}} \\
    \midrule
    n-shot & \multicolumn{2}{c}{F1(SUPPORTS)} & \multicolumn{2}{c}{F1(NOT\_ENOUGH\_INFO)} & \multicolumn{2}{c}{F1(REFUTES)} \\
     &        mean &    std &               mean &    std &       mean &    std \\
    \midrule
1  &      0.3333 & 0.1540 &             0.3325 & 0.1679 &     0.3169 & 0.1363 \\
2  &      0.3750 & 0.1367 &             0.3810 & 0.1415 &     0.3541 & 0.1191 \\
3  &      0.4218 & 0.1159 &             0.4099 & 0.1263 &     0.3663 & 0.0926 \\
4  &      0.4162 & 0.1119 &             0.4299 & 0.1154 &     0.3805 & 0.0885 \\
5  &      0.4251 & 0.1044 &             0.4538 & 0.1005 &     0.3836 & 0.0773 \\
    \bottomrule
        \toprule
    \multicolumn{7}{c}{\textbf{SciFact\_oracle}} \\
    \midrule
    n-shot & \multicolumn{2}{c}{F1(SUPPORTS)} & \multicolumn{2}{c}{F1(NOT\_ENOUGH\_INFO)} & \multicolumn{2}{c}{F1(REFUTES)} \\
     &        mean &    std &               mean &    std &       mean &    std \\
    \midrule
1  &      0.3326 & 0.1764 &             0.5141 & 0.1518 &     0.3346 & 0.1568 \\
2  &      0.3295 & 0.1326 &             0.5702 & 0.1192 &     0.3794 & 0.0961 \\
3  &      0.3780 & 0.1168 &             0.5931 & 0.0741 &     0.3750 & 0.0766 \\
4  &      0.3849 & 0.1090 &             0.5882 & 0.0879 &     0.3830 & 0.0737 \\
5  &      0.3975 & 0.0992 &             0.5943 & 0.0656 &     0.3744 & 0.0746 \\
    \bottomrule
        \toprule
    \multicolumn{7}{c}{\textbf{SciFact\_retrieved}} \\
    \midrule
    n-shot & \multicolumn{2}{c}{F1(SUPPORTS)} & \multicolumn{2}{c}{F1(NOT\_ENOUGH\_INFO)} & \multicolumn{2}{c}{F1(REFUTES)} \\
     &        mean &    std &               mean &    std &       mean &    std \\
    \midrule
1  &      0.3369 & 0.1542 &             0.5438 & 0.1751 &     0.3519 & 0.1525 \\
2  &      0.3612 & 0.1199 &             0.5910 & 0.1524 &     0.3930 & 0.1117 \\
3  &      0.4030 & 0.0983 &             0.6407 & 0.1045 &     0.3868 & 0.0949 \\
4  &      0.4063 & 0.0822 &             0.6409 & 0.0857 &     0.3859 & 0.0922 \\
5  &      0.3994 & 0.0867 &             0.6555 & 0.0632 &     0.3989 & 0.0713 \\
    \bottomrule
    \end{tabular}
    \caption{MAPLE Classwise F1 results. The reported results are mean and standard deviation classwise F1 scores for each class on 100 runs.}
    \label{tab:classwise}
\end{table*}

\subsection{Performance comparison within 50 shots}
\label{appendix: performance within 50 shots}

\begin{figure*}[htb]
    \centering
    \includegraphics[scale=0.28]{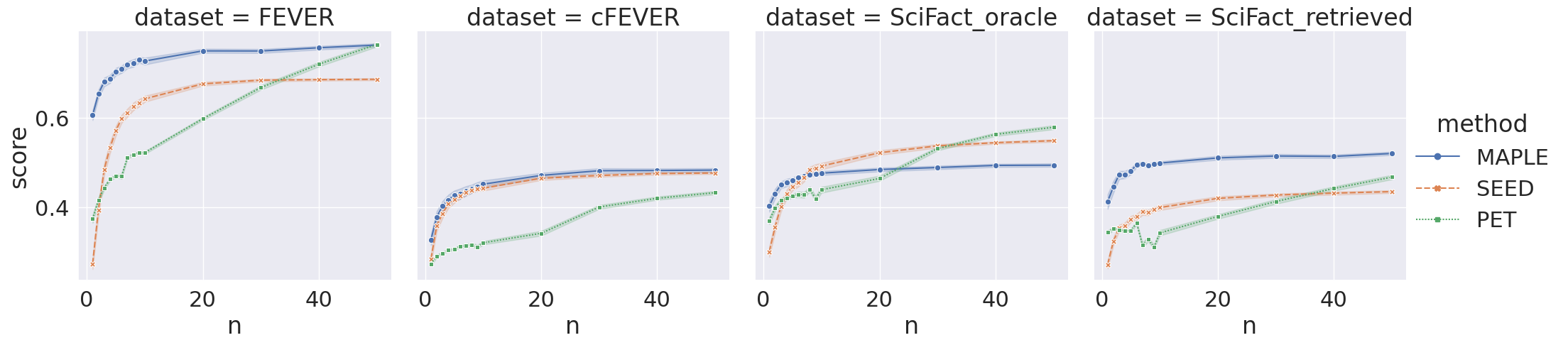}
    \caption{F1 performance within 50 shots.}
    \label{fig:max50}
\end{figure*}

Figure \ref{fig:max50} illustrates the F1 results within the 50-shot setting. The experiments are conducted on SEED, PET and MAPLE, as LLaMA 2 imposes high demand on computational budget. MAPLE demonstrates superior performance in three out of four dataset configurations, specifically FEVER, cFEVER, and SciFact\_retrieved. Although it is not the top performing approach in the SciFact\_oracle setting, it holds the highest position until surpassed by SEED at 8 shots, followed by PET at 30 shots.

On the FEVER dataset, MAPLE achieves significant improvements over the baselines when provided with fewer than 50 shots. MAPLE starts with a very high performance around 0.6 and converges around 20 shots, reaching approximately 0.8. Despite starting from a very low point, SEED learns rapidly within 10 shots and converges around 20 shots with a score below 0.7. PET demonstrates remarkable learning capabilities within 50 shots, as its performance steadily rises to around 0.8.

On the cFEVER dataset, MAPLE remains the best-performing method within 50 shots, although with only a slight margin over SEED. Both MAPLE and SEED exhibit similar performance curves, converging around 20 to 30 shots with scores approaching 0.5. PET shows a different pattern, steadily learning over the range of 50 shots but ending with a lower score compared to the other methods.

On the SciFact\_oracle dataset, MAPLE starts strongly but shows limited improvements with more data, converging within 8 shots at approximately 0.48. This may be attributed to the challenging nature of the scientific domain. SEED and PET manage to surpass MAPLE in this case, with SEED converging at 50 shots and achieving a score of around 0.55. PET surpasses MAPLE after being provided with over 20 shots and surpasses SEED after receiving over 30 shots.

On the SciFact\_retrieved dataset, unlike in the SciFact\_oracle case, MAPLE maintains a clear advantage within 50 shots. MAPLE starts above 0.4 and converges around  20 to 30 shots with a score above 0.5. With retrieved evidence, both SEED and PET experience a performance dip compared to the oracle evidence scenario. SEED also converges around 20 to 30 shots, but with a score above 0.4. PET experiences a dip early on, around 10 shots, dropping to approximately 0.3, despite starting around 0.35. Afterwards, it recovers and reaches above 0.45 at 50 shots, although still lower than MAPLE.


\section{Runtime Appendix}
\subsection{LoRA vs SFT Runtime comparison}
\label{appendix: LoRA vs SFT runtime}

\begin{table*}[htbp]
\small
\centering
\begin{tabular}{lcccc}
\hline
\textbf{ } & \textbf{FEVER} & \textbf{cFEVER} & \textbf{SciFact\_oracle} & \textbf{SciFact\_retrieved} \\
\hline
\textbf{LoRA runtime (from claim to evidence)} & 00:50:24 & 00:39:14 & 00:05:33 & 00:16:29 \\
\hline
\textbf{SFT runtime (from claim to evidence)} & 01:50:52 & 01:15:14 & 00:13:23 & 00:48:21 \\
\hline
\textbf{LoRA runtime (from evidence to claim)} & 00:50:23 & 00:39:12 & 00:05:18 & 00:16:28 \\
\hline
\textbf{SFT runtime (from evidence to claim)} & 01:37:58 & 01:14:39 & 00:11:41 & 00:35:12 \\
\hline
\end{tabular}
\caption{LoRA vs SFT Runtime comparison. The time format is hours:minutes:seconds.}
\label{tab:LoRA runtime}
\end{table*}

We present the runtime comparison of LoRA and SFT on performing Seq2seq training on T5-small. While the efficiency gain varies on the given training data, table \ref{tab:LoRA runtime} shows that significant time savings across all experimented datasets.

\subsection{Overall Runtime}
\label{appendix: overall runtime}
We present the runtime of MAPLE across four dataset configurations in Table \ref{tab:overall runtime}. The experiments were conducted on a High-Performance Compute cluster provided by the university, featuring 8 compute cores, 11G RAM per core, and a single NVIDIA A100 GPU. Seq2seq LoRA training and SemSim transformation were applied to the entire dataset. The LR runtime denotes the execution time for all few-shot experiments outlined in Section \ref{experiments}. It's important to note that the runtime is strongly correlated with the size of the unlabelled pool, as well as the length of claims and evidences. Consequently, it takes a few hours to run for large-scale datasets like FEVER and cFEVER, as well as dataset configurations comprising lengthy instances such as SciFact\_retrieved, but considerably less time for SciFact\_oracle. For improved efficiency, future work may explore applying the SemSim transformation solely to the sampled few-shot training instances per experiment.

\begin{table*}[htbp]
\small
\centering
\begin{tabular}{lcccc}
\hline
\textbf{ } & \textbf{FEVER} & \textbf{cFEVER} & \textbf{SciFact\_oracle} & \textbf{SciFact\_retrieved} \\
\hline
\textbf{Seq2Seq runtime (from claim to evidence)} & 00:50:24 & 00:39:14 & 00:05:33 & 00:16:29 \\
\hline
\textbf{SemSim runtime (from claim to evidence)} & 00:50:16 & 00:37:34 & 00:06:22 & 00:26:06 \\
\hline
\textbf{Seq2Seq runtime (from evidence to claim)} & 00:50:23 & 00:39:12 & 00:05:18 & 00:16:28 \\
\hline
\textbf{SemSim runtime (from evidence to claim)} & 00:49:02 & 00:37:34 & 00:05:45 & 00:23:06 \\
\hline
\textbf{LR runtime} & 00:00:28 & 00:00:33 & 00:00:31 & 00:00:33 \\
\hline
\textbf{Total runtime} & 03:20:33 & 02:34:07 & 00:23:29 & 01:22:42 \\
\hline
\end{tabular}
\caption{MAPLE runtime on four dataset configurations. The time format is hours:minutes:seconds.}
\label{tab:overall runtime}
\end{table*}

\end{document}